\documentclass{article}
\usepackage{ijcai22}

\usepackage{times}

\usepackage{soul}
\usepackage{url}
\usepackage[hidelinks]{hyperref}
\usepackage[utf8]{inputenc}
\usepackage[small]{caption}
\usepackage{graphicx}
\usepackage{amsmath}
\usepackage{booktabs}
\usepackage{algorithmic}
\usepackage{amssymb}
\usepackage{bbm}
\usepackage{xcolor}

\usepackage{times}
\usepackage{epsfig}
\usepackage{graphicx}
\usepackage{amsmath}
\usepackage{amssymb}
\usepackage{booktabs}

\usepackage{times}
\usepackage{epsfig}
\usepackage{graphicx}

\usepackage{multirow}
\usepackage{blindtext}
\usepackage{placeins}
\usepackage{adjustbox}
\usepackage{xfrac}
\usepackage{enumitem}
\usepackage{caption}
\usepackage[linesnumbered,boxed,lined]{algorithm2e}

\usepackage{array}
\usepackage{rotating}
\usepackage{multicol}
\usepackage{chngcntr}
\usepackage{bm}
\usepackage{booktabs}

\usepackage{comment}
\usepackage{lipsum}
\usepackage{subfig}

\newcommand{\approach}{GCRN~}

\usepackage{ifthen}
\newboolean{combined}
\setboolean{combined}{true}

\newcommand{\beginsupplement}{%
        \setcounter{table}{0}
        \renewcommand{\thetable}{S\arabic{table}}%
        \setcounter{figure}{0}
        \renewcommand{\thefigure}{S\arabic{figure}}%
        
        \setcounter{section}{0}
        \renewcommand{\thesection}{S\arabic{section}}%
        
        \setcounter{page}{1}
        \renewcommand{\thepage}{S\arabic{page}}%
}

\title{Detecting out-of-context objects using contextual cues}

\author{
Manoj Acharya$^1$\and Anirban Roy$^2$\and
Kaushik Koneripalli$^{2}$\and
Susmit Jha$^{2}$\and
Christopher Kanan$^{1,3,4}$\And
Ajay Divakaran$^2$\\
\affiliations
$^1$Rochester Institute of Technology, Rochester NY 14623, USA\\
$^2$SRI International,Princeton NJ 08540, USA\\
$^3$Paige, New York NY 10036, USA \and
$^4$Cornell Tech, New York NY 10044, USA\\
\emails
\{manoj,kanan\}@rit.edu,
\{anirban.roy, kaushik.koneripalli,
susmit.jha, ajay.divakaran\}@sri.com
}

\begin{document}
\maketitle
\begin{abstract}
This paper presents an approach to detect out-of-context (OOC) objects in an image. Given an image with a set of objects, our goal is to determine if an object is inconsistent with the scene context and detect the OOC object with a bounding box. In this work, we consider commonly explored contextual relations such as co-occurrence relations, the relative size of an object with respect to other objects, and the position of the object in the scene. We posit that contextual cues are useful to determine object labels for in-context objects and inconsistent context cues are detrimental to determining object labels for out-of-context objects. To realize this hypothesis, we propose a graph contextual reasoning network (GCRN) to detect OOC objects. GCRN consists of two separate graphs to predict object labels based on the contextual cues in the image: 1) a representation graph to learn object features based on the neighboring objects and 2) a context graph to explicitly capture contextual cues from the neighboring objects. GCRN explicitly captures the contextual cues to improve the detection of in-context objects and identify objects that violate contextual relations. 
In order to evaluate our approach, we create a large-scale dataset by adding OOC object instances to the COCO images. We also evaluate on recent OCD benchmark. Our results show that GCRN outperforms competitive baselines in detecting OOC objects and correctly detecting in-context objects.
\end{abstract}

\section{Introduction}
\label{sec:intro}

Our goal is to detect objects that appear out-of-context (OOC) in an image. Given an image with a set of objects, the task we tackle is to detect objects that are OOC and also correctly identify the in-context objects. Typically, objects in natural images appear in a suitable context and it is useful to consider context while detecting objects \cite{choi_cvpr10,oliva2007role,gould2008multi,divvala2009empirical,koller2009probabilistic,beery2018recognition,zhang2012efficient,sun2017seeing,bomatter2021pigs}. While appropriate contextual cues are shown to be useful for object detection, incorrect contextual cues can negatively impact the performance of object detection for both humans \cite{puttingzhang2020,bomatter2021pigs} and machine learning approaches \cite{torralba2003contextual,choi_cvpr10,rosenfeld2018elephant,madras2021identifying}. Thus, in order to develop reliable object detection systems, it is crucial to detect objects that appear in unusual contexts where the predictions may not be reliable. 

\begin{figure}[t]
\begin{center}
\includegraphics[width=\linewidth]{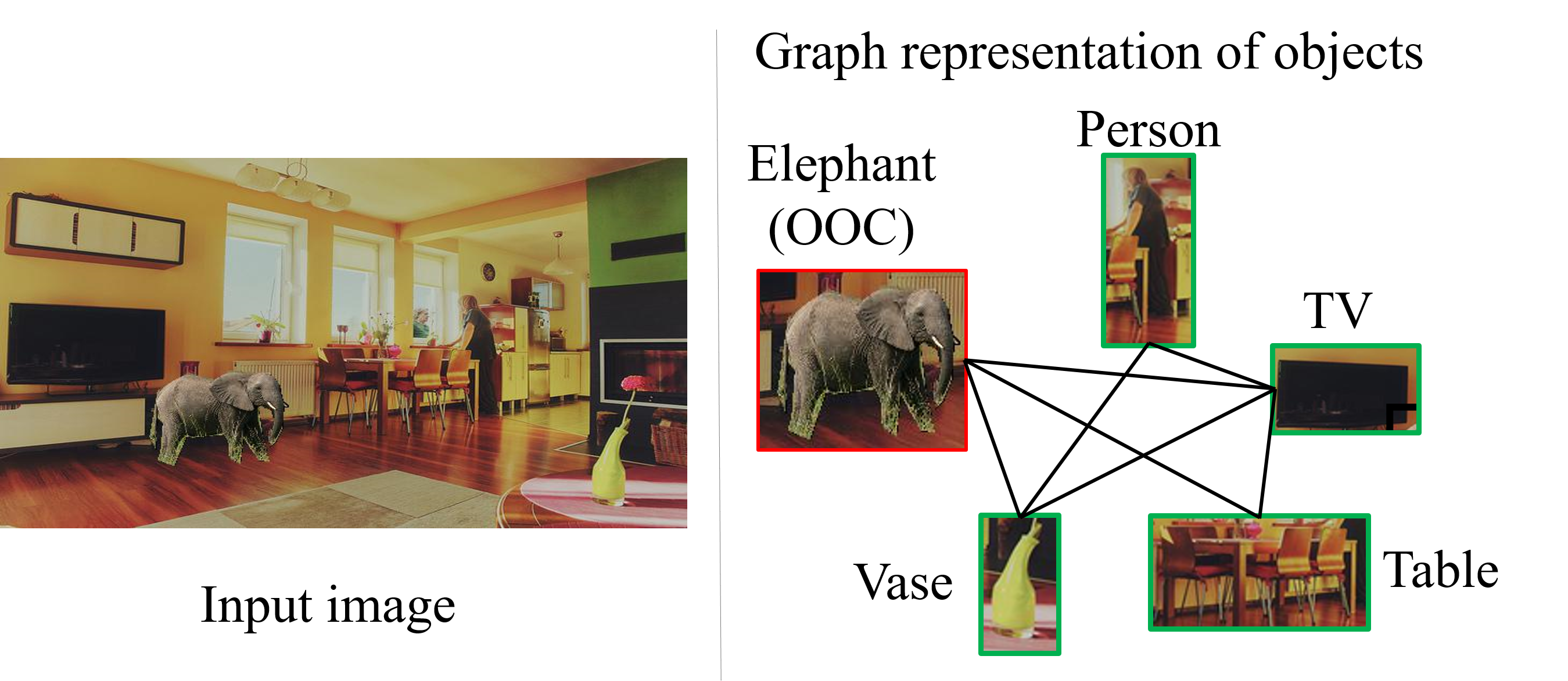}
\end{center}
\caption{We aim to detect OOC object instances in an image. Given the `elephant in the room' image, we build a graph based on the objects to share context cues between the objects. In-context objects are highlighted with green frames and the OOC object is highlighted with a red frame. Here, the `elephant'  is OOC as it is not consistent with the co-occurrence relations with other objects in the image.}
\label{fig:intro}
\end{figure}

While detecting in-context objects is extensively explored, detecting OOC objects is much less well studied; however, recent work has shown that the presence of OOC objects can severely affect an object detectors' ability to detect in-context objects in the image  \cite{rosenfeld2018elephant}. Detecting OOC objects is difficult for humans as well \cite{bomatter2021pigs}. Motivated by these observations, we develop an approach for detecting OOC objects by capturing contextual cues and checking inconsistent object-context relations.


Detecting OOC objects is challenging as these objects appear normal in isolation and it is only the context that makes them unusual. We also assume that OOC object classes are present during training and thus relying on object appearance alone may not be sufficient for OOC detection. A few early works aim to detect OOC objects \cite{choi_cvpr10} where hand-crafted contextual relationships (e.g., co-occurrence, the constraint on object size) are exploited for OOC detection. Some recent approaches consider neural network-based models to learn contextual relations in a data-driven manner \cite{bomatter2021pigs,dvornik2018modeling}. Many of such approaches model context as generic background and do not exploit informative cues such as label dependencies and objects properties in the image \cite{beery2018recognition,puttingzhang2020}. To address these challenges, we propose an approach to explicitly consider contextual cues for object detection in order to discover the inconsistency for OOC detection. We posit a simple yet effective hypothesis to detect OOC objects in an image - for in-context objects, label predictions with and without contextual cues are expected to match, and for out-of-context objects, predictions with and without context are \emph{not} expected to match as the objects are not consistent with the context. Contextual cues are only useful to detect objects that are in-context and adversely affect the detection of out-of-context objects. For example, as shown in Fig.~\ref{fig:intro}, detecting in-context objects such as `person' can benefit from the context cues in terms of other indoor objects. However, in the case of the OOC `elephant' object, inconsistent contextual cues can be confusing as elephants usually do not appear in indoor scenes.

We propose a graph contextual reasoning network (GCRN) to capture contextual cues for predicting object class prediction. We model context by defining a graph over the objects in an image where nodes represent objects and edges represent object-to-object relations. Specifically, our GCRN model consists of two graph models: 1) representation graph (repG) to learn useful node representations to predict object labels, and 2) context graph (conG) to capture contextual cues to predict object labels. We consider three contextual cues for OOC detection: co-occurrence of objects, location, and shape similarity of object boxes. These contextual cues are shown to effective for in-context object detection \cite{choi_cvpr10,gould2008multi,divvala2009empirical,koller2009probabilistic,zhang2012efficient}. Each graph model is realized by a Graph Convolutional Network (GCN) to ensure efficient learning and inference \cite{dai2016discriminative,kipf2016semi,hamilton2017inductive,velickovic2019deep,qu2019gmnn}. Both the models are trained together to ensure GCRN learns informative node representation as well as context-dependency among the objects. GCRN has several advantages over existing approaches. Compared to the graph-based models that define contextual relations using hand-crafted features, such as conditional random fields \cite{gould2008multi,gould2009decomposing,zhang2012efficient}, \approach learns these cues in a data-driven manner. Compared to the standard GCN models that do not consider the dependency among the node predictions \cite{dai2016discriminative,kipf2016semi,hamilton2017inductive}, conG in GRCN explicitly captures the context while predicting node labels. Our experiments show GCRN significantly outperforms standard GCNs in OOC detection.

Our main contributions include:

\begin{itemize}
    \item We propose an approach to simultaneously detect OOC objects and in-context objects in an image.
    \item We propose a graph contextual reasoning network (GCRN) to detect OOC objects by explicitly modeling contextual cues to predict object labels. GCRN allows a flexible framework to model and learn contextual cues in a data-driven manner. Unlike prior works, GCRN does not require manual specification of the contextual constraints.
    \item We create a large-scale OOC dataset to evaluate our approach. The dataset consists of roughly 100K OOC images where objects occur in various OOC scenarios. 
\end{itemize}

\section{Related Works}
\label{sec:rel_work}

\textbf{Context for object detection.} 
Contextual cues are important for object detection and segmentation. Graph-based models provide a flexible way to represent context where nodes represent objects and edges represent pair-wise relations among the objects \cite{choi_cvpr10,gould2008multi,zhang2012efficient}. Among graph-based models, conditional random fields (CRF) are explored extensively where contextual cues are represented by edge potentials. Common contextual cues include co-occurrence, spatial distance, geometric and appearance similarity \cite{choi_cvpr10,gould2008multi,divvala2009empirical,koller2009probabilistic,zhang2012efficient}. More recently, graphical models are combined with neural networks to exploit data-driven feature learning. Graph convolutional networks (GCN) \cite{dai2016discriminative,kipf2016semi,hamilton2017inductive} provide a convolutional implementation of the graphical models combining the power of representation learning of neural networks with the structured representation of graphs. However, standard GCNs do not explicitly capture the contextual relations that are crucial to detect OOC objects \cite{qu2019gmnn}. Our \approach learns two GCNs, one for learning the feature representation and another for capturing context cues, to effectively detect OOC objects.

\textbf{Out-of-context object detection.}
Existing studies have argued the importance of OOC object detection as these affect the performance of object detection for both humans and machines \cite{choi_cvpr10,rosenfeld2018elephant,puttingzhang2020,bomatter2021pigs}. Choi et al. \cite{choi2012context} define OOC objects that violate common contextual rules (e.g., flying cars) in terms of unusual background, unusual size, etc. They consider a graph-based model to capture such relations among the objects. Some of the approaches consider context as the entire background with respect to an object and detect OOC objects that are inconsistent with the background. Dvornik et al. \cite{dvornik2018modeling} usual pair of objects as OOC instances. Unlike these approaches, we define the context for a target object as its relation (object classes, relative size, relative location) with other objects and the scene.

\textbf{Graph Convolutional Networks.}
GCNs provide an end-to-end neural network-based realization of graph models and are shown to be successful in object detection \cite{dai2016discriminative,kipf2016semi,hamilton2017inductive,qu2019gmnn}. In GCNs, a robust node representation is learned by sharing the representations with neighboring nodes. This node-to-node exchange is implemented via a convolutional operation that facilitates efficient learning and inference in GCNs. However, GCNs typically avoid modeling the label dependency among the nodes \cite{qu2019gmnn}. Thus, common GCNs frameworks are not effective to capture global contextual cues. Recently, graph Markov neural networks (GMNN) \cite{qu2019gmnn} are proposed to capture label dependency by simultaneously learning two GCNs - one for learning node representations and another for learning label dependency. We consider a similar setup where one network learns the node representation and another network captures context for predicting mode labels.

\section{Proposed approach}
\label{sec:approach}

\begin{figure*}[t]
\begin{center}
\includegraphics[width=0.8 \linewidth]{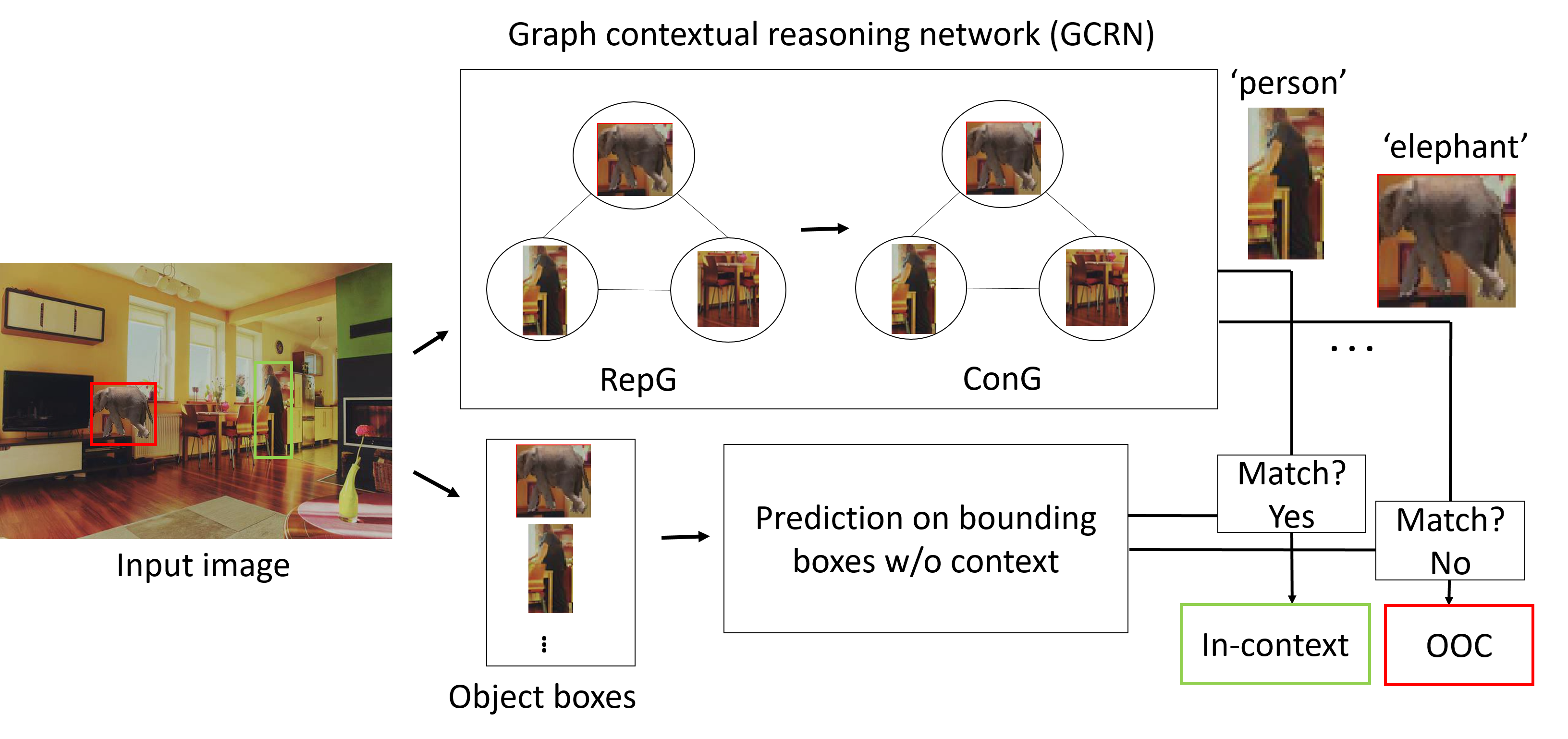}
\end{center}
\caption{GCRN framework takes in an input image with bounding boxes to construct an image graph G~=~(V, E) with vertices V and edges E. The graph G is fed into a RepG network that learns object dependent representation and also ConG network that learns the contextual dependency of objects. Finally, for any vertex $v$ from V in relation to graph G, we compare with the learned context to determine if it is Out-Of-Context (OOC).}
\label{fig:framework}
\end{figure*}

In order to detect out-of-context (OOC) objects in an image, we hypothesize that, for OOC objects, accurate labels cannot be predicted from the context as these objects are inconsistent with the context. On the other hand, contextual cues are expected to help predict the labels for in-context objects \cite{choi_cvpr10,oliva2007role,gould2008multi,gould2009decomposing,divvala2009empirical,koller2009probabilistic,beery2018recognition,zhang2012efficient}. To realize this hypothesis, we propose a graph contextual reasoning network (GCRN) that consists of two graph models: 1) Representation graph (RepG) that learns precise object representation at each node by sharing the representations with its neighbors. RepG relies on the shared representations to predict object labels ignoring context dependency among the labels \cite{qu2019gmnn}. 2) To complement RepG, we propose a context graph (ConG) that learns context dependency at each node by sharing the context cues with its neighbors. We consider graph convolutional networks (GCN) \cite{kipf2016semi,hamilton2017inductive} to instantiate both RepG and ConG. We first introduce the GCN framework and then discuss the implementation details of RepG and ConG. 

\textbf{Graph convolutional network.} Given an image, we build a graph over the objects. Consider a graph $G = (V, E)$, where $V$ is the set of nodes and $E$ is the set of edges. We define $X = \{ \bm{x}_i \}$  and $Y = \{ y_i \}$  as the feature representation and label of $i$th node, respectively. Given this definition, the goal is to predict object labels from the features. 
\begin{equation}
    H^{l+1} = f^l(W^l, H^l, E), \: p(y_i | X, E) = \text{SM}(H^L),
    \label{equ:gcn}
\end{equation}
where $f^l(\cdot)$ is the convolutional function corresponding to a layer $l$. $f^l(\cdot)$ iteratively updates node representations to $H^{l+1}$ from the current representation $H^l$ and using the edges between nodes $E$. $W^l$ represents the parameter for layer $l$. Note that $H^0 = X$, the initial node representations. The prediction is made by applying a softmax operation ($\text{SM}$) on the final representation $H^L$.

\textbf{Context-informed prediction.} We formulate context-informed label prediction in a conditional random field framework \cite{lafferty2001conditional,koller2009probabilistic} where the conditional distribution over the labels is given by
\begin{equation}
    p(Y | X, E) = \frac{1}{Z(X, E)} \prod_{(i, j) \in E} \phi_{i,i}(y_i, y_j, X),
    \label{equ:RF}
\end{equation}
where $Z(X, E)$ is the partition function over the graph and $\phi_{i,i}(y_i, y_j, X )$ is the potential function over a pair of nodes $i, j$ \cite{gould2008multi,gould2009decomposing,koller2009probabilistic}. Lets denote $\theta$ as the graph parameters. Then, learning can be done by maximizing the following conditional log-likelihood:
\begin{equation}
    \ell_{Y|X} (\theta) = \text{log} \: p_{\theta}(Y | X, \theta).
    \label{equ:likelihood}
\end{equation}
However, directly maximizing this likelihood function is intractable due to the combinatorial nature of the partition function \cite{koller2009probabilistic}. Thus, we consider optimizing the approximate likelihood that is shown to successful in learning similar graphical models \cite{richardson2006markov}.
\begin{equation}
    \ell_{Y|X} (\theta) = \sum_{i} \text{log} \: p_{\theta}(y_i | y_{j \in B(i)}, X, \theta),
    \label{equ:likelihood}
\end{equation}
where $B(i)$ is the set of neighbors of $i$. Intuitively, with this approximation, we only consider the context dependencies based on the neighboring nodes. However, as we consider GCN to capture this dependencies, a few iterations of GCN allows to capture log-range dependencies by iterative message passing \cite{koller2009probabilistic}. Note that even in Equ.~\ref{equ:likelihood}, the label of a node $y_i$ is conditioned on both the neighboring context ($y_{j \in B(i)}$) and the feature representation $X$. To avoid this dependency, we propose an iterative optimization where in one phase, we optimize to learn only representations and the context dependency in the following phase. This allows each iteration to be optimized efficiently in a pair of GCN framework.

\textbf{Representation graph.} In the first phase, we consider a mean-field approximation to remove context dependency and learn only the representation. Thus, the label distribution is defined as
\begin{equation}
    p_{\theta_R}(Y | X, E) = \prod_{i} p_{\theta_R}(y_i | X, E),
    \label{equ:mena_field}
\end{equation}
where $\theta_R$ is the parameters for representation graph. Our representation graph is implemented by a GCN where object features are used as node representations. 
\begin{equation}
    H^{l+1}_R = f^l_R(W^l_R, H^l_R, E), \: p(y_i | X_R, E) = \text{SM}(H^L_R),
    \label{equ:R_gcn}
\end{equation}
where $X_{R}$ denote the node representations and $W_R$ denotes the GCN parameters. We consider bounding box features as the node representations. It is evident from Equ. \ref{equ:gcn} and Equ. \ref{equ:mena_field} that neighboring nodes share feature representations by message passing but context related to object labels are not shared.

\textbf{Context graph.} In the second phase, we aim to predict labels from the context. This is also realized by a context GCN as follows.
\begin{equation}
    H^{l+1}_C = f^l_C(W^l_C, H^l_C, E), \: p(y_i | X_C, E) = \text{SM}(H^L_C),
    \label{equ:R_gcn}
\end{equation}
where $X_{C}$ denote the context-related node representations and $W_C$ denotes the GCN parameters. We consider context features as the node representations including class labels, position, size of neighboring object boxes. 

\textbf{Learning.} We perform an iterative learning in an expectation-maximizing (EM) framework \cite{richardson2006markov,qu2019gmnn}. In the M-step, we learn the representation graph based on the predicted labels of the context graph and in the E-step, keeping the representation fixed, we update the context graph based on the ground truth labels. Specifically, in E-step, we start with a pre-trained RepG to predict node labels. Then the labels are updated by a fixed ConG by aggregating context from the neighboring nodes. Finally. the RepG is updated to match ConG's predictions. In M-step, we update ConG based on the ground-truth labels. We continue these iterations until convergence, i.e., the difference of the predictions between RepG and ConG becomes zero or below a threshold. 

\textbf{Detecting OOC instances.} Given an image, we localize all objects by bounding boxes and consider them as the nodes of the graph. GCRN predicts softmax distribution over class labels for each box by considering context cues from other boxes. We also train an object classifier to predict object labels only from the bounding box ignoring the context cues. Finally, we compare the KL divergence (KLD) between the softmax distribution from both predictions as a measure of OOC. In in-context objects, KLD is expected to be low and for OOC objects, the KLD is expected to be high. Thus we can detect OOC instances by choosing an appropriate threshold. Our prediction framework is shown in figure~\ref{fig:framework}.

\textbf{Implementation Details.}
We implement the GCRN framework using the DGL toolbox~\cite{wang2019deep}. For both RepG and ConG, we consider GCN with four graph convolution layers with residual connections between the layers. The numbers of neurons at these layers are 256, 128, 64, and 64 respectively. Models are trained using an AdamW optimizer with a learning rate of 0.001 without decay. For our GCRN framework, we first train RepG for five epochs in the first phase and alternate between RepG and ConG until convergence. In our experiments, convergence is reached within ten iterations. Residual connections between the layers were crucial for efficient learning.

To create a graph for an image, we detect objects in images and then extract features for the objects to seed the graph nodes. We consider a MaskRCNN \cite{he2017mask} model pre-trained on the COCO dataset to detect objects. We train a ResNet50 \cite{he2016deep} network for feature feature extraction. For context cues, we consider additional geometric features from the bounding box as $(xmin, ymin, xmax,ymax)$ and 7D spatial feature vector $\left[\frac{w}{W} , \frac{h}{H} ,\frac{a}{A}, \frac{xmin}{W} , \frac{ymin}{H} , \frac{xmax}{W}, \frac{ymax}{H}\right]$ where $w$, $h$, $a$ represent the width, height and area of the bounding box and $W$, $H$, $A$ represent the same for the image. These spatial features are important to capture usual size and location of objects training images. 

\section{Experiments}
\label{sec:experimets}
In the following, we introduce the dataset, describe experimental setup, metrics, and present results.

\textbf{Datasets.}

\textbf{The COCO-OOC Dataset.} We create a large-scale OOC dataset to evaluate OOC detection due to the absence of such datasets. As we consider objects and their properties as context, we require object annotations for all the objects along with the OOC object. We consider COCO 2014~\cite{lin2014microsoft} dataset, consisting 80 indoor and outdoor objects classes, to create the COCO-OOC dataset. Following the common strategy \cite{blum2021fishyscapes,puttingzhang2020}, we place objects in images that violate the contextual relations. We leverage available object mask to transplant objects in images to create OOC scenarios. COCO-OOC consists of 106,036 images with three types of OOC  violating co-occurrence relation and size constraints \cite{blum2021fishyscapes,puttingzhang2020,bomatter2021pigs}. Additional details about OOC creation and example images are provided in the supplemental material.

\textbf{The OCD Dataset.}
In addition to our own dataset, we also use the recent OCD benchmark that has synthetically generated OOC indoor scenes \cite{bomatter2021pigs}. The OOC images are generated from the VirutalHome \cite{puig2018virtualhome} environment. OCD has 11,155 OOC images where objects violate co-occurrence, size, and gravity constraints. 

\textbf{Experimental setup and metrics.} We train the GCRN model on the COCO train images where objects appear in-context to capture the usual context for object. We test on OOC images aiming to detect the OOC object instance. Note that we assume all OOC objects are available in training and are not novel during the test. Thus relying on appearance cues is not sufficient to detect the OOC instances. As selecting a threshold is often crucial to separating OOC from in-context objects, we consider the AUC score as the metric to robustly evaluate our performance.

\textbf{Baselines.} We propose the following baselines based on the state-of-the-art approaches to evaluate various aspects of our \approach. 

\textbf{Softmax confidence.} In this baseline, we consider the softmax confidence as a measure of OOC assuming that the confidence would be lower for the OOC objects than usual in-context objects. Softmax confidence is successfully used to detect anomaly \cite{blum2021fishyscapes} and novel objects \cite{liang2017enhancing}. The results in table~\ref{table:basline} imply the softmax confidence is not reliable to detect OOCs as, unlike anomalous or novel objects, the OOC objects are observed during training. This baseline implies that relying on objects' appearance is not sufficient to detect OOC objects.

\textbf{Without context graph.} In this baseline, we do not explicitly capture context. Specifically, we omit the context graph and only consider the representation graph. The results in table~\ref{table:basline} imply that context graph is important to capture context and representation graph itself is not sufficient to capture context dependencies. This baseline is comparable with the \cite{bomatter2021pigs} where object representation is learned through a shared network and context dependencies between the objects (e.g., label dependencies) are not modeled. 

\begin{table}
\caption{Comparison with the baselines on the COCO-OOC dataset.}
\label{table:basline}
\centering
\begin{tabular}{lc}
\toprule
{Approach}  & {AUC score} \\
\midrule
Softmax confidence & 0.043
\\
GCRN (w/o ConG) & 0.589
\\
GCRN & 0.980
\\
\bottomrule
\end{tabular}
\end{table}

\textbf{Impact of context on In-context vs. OOC object detection.} We compare the performance of object detection for in-context and OOC objects on COCO-OOC dataset. The results are shown in table~\ref{table:accuracy}. As expected, the performance on OOC objects is inferior. Note that ignoring explicit context cues (w/o ConG) degraded the performance for both in-context and OOC objects. This justifies our hypothesis that accurate contextual are helpful for object detection while inconsistent context cues can be confusing.

\begin{table}
\caption {Accuracy score across OOC and non-OOC instances in our COCO-OOC dataset. For OOC instances, miclassification refers to correct detection, thus lower accuracy is better while for non-OOC instances higher is better.}
\label{table:accuracy}
\centering
\begin{tabular}{lcccc}
\toprule
Approach &  OOC$\downarrow$ & non-OOC$\uparrow$ & Overall  \\
\midrule
GCRN (w/o ConG) & 0.69 & 0.77 & 0.76 \\
GCRN            & 0.30 & 0.98 & 0.93  \\
\bottomrule
\end{tabular}
\end{table}

\textbf{Types of OOC.} We evaluate \approach's performance on detecting OOC objects that violate co-occurrence and size constraints in table~\ref{table:ooc_variants}. The performance on co-occurrence is superior as detecting the violation of co-occurrence is relatively easier. For size OOC, we consider the relative size of the object boxes to capture these contexts. However, in 2D images, actual size information is partially lost due to the camera projection which makes it harder to capture size context.

\begin{table}
\caption{AUC results for detecting OOC variants on COCO-OOC.}
\label{table:ooc_variants}
\centering
\begin{tabular}{ccc}
\toprule
OOC type  & Co-occurrence & Size 
\\
\midrule
AUC score & 0.986  & 0.921
\\
\bottomrule
\end{tabular}
\end{table}

\textbf{Impact of object detector on OOC detector.} It is important to detect an object to determine whether it is an OOC instance. In this section, we analyze the effect of object detection on OOC detection performance. We consider three modes of operation for our approach: 1) oracle bounding boxes with labels (oracle boxes+labels) where we assume to have access to object locations in the form of bounding boxes and object labels, 2) oracle bounding boxes (oracle boxes, pred labels) where we assume to have to access to only the object locations in the form of bounding boxes but objects labels are to be predicted, 3) predicted bounding boxes (pred boxes) where we do not have access to bounding boxes and an object detector is used to discover the objects. In the oracle boxes+labels setup, errors are completely attributed to OOC detection. In oracle boxes, additional errors may come from misidentifying object class from object classifier. In pred boxes, errors may come from incorrect boxes or wrong labels. The results are shown in table~\ref{table:object_detector_error}. As expected, the performance gradually decreases from oracle boxes+labels to pred boxes. We consider state-of-the-art approaches for both object detection and object classification but these can be further improved to achieve a better performance in OOC detection.

\begin{table}
\caption{Impact of the choice of bounding box detector on OOC detection for the COCO-OOC dataset.}
\label{table:object_detector_error}
\centering
\begin{tabular}{lc}
\toprule
{Approach}  & {AUC score} \\
\midrule
GCRN (oracle boxes + labels) & 0.980 
\\
GCRN (oracle boxes, pred labels) & 0.897
\\
GCRN (pred boxes) & 0.771  
\\
\bottomrule
\end{tabular}
\end{table}

\begin{figure*}[t]
\centering
\begin{tabular}{lll}
\includegraphics[width=0.28\linewidth]{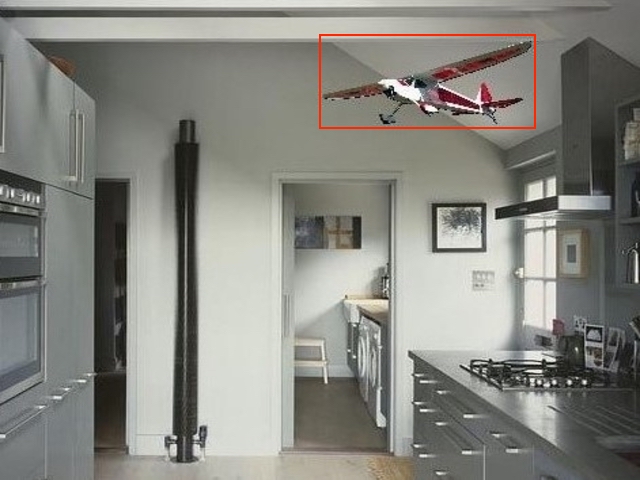}
&
\includegraphics[width=0.28\linewidth]{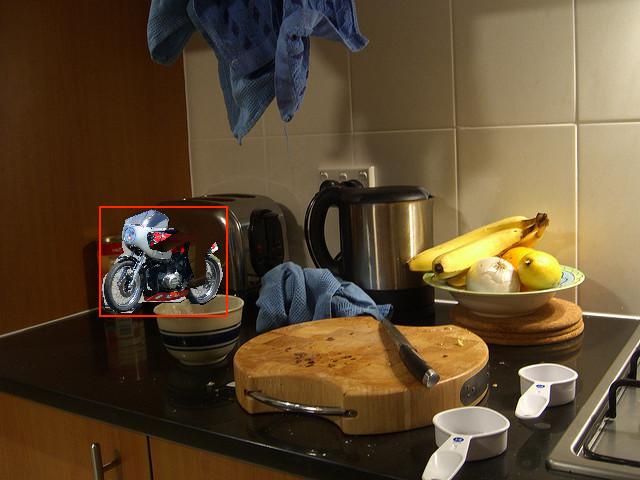}
&
\includegraphics[width=0.28\linewidth]{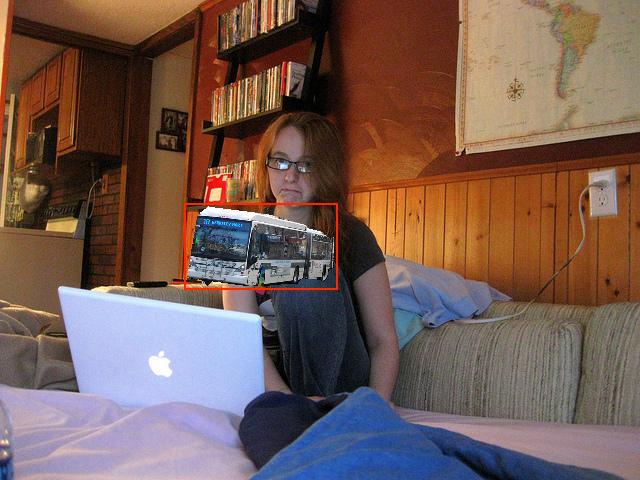}
\\\\
\includegraphics[width=0.28\linewidth]{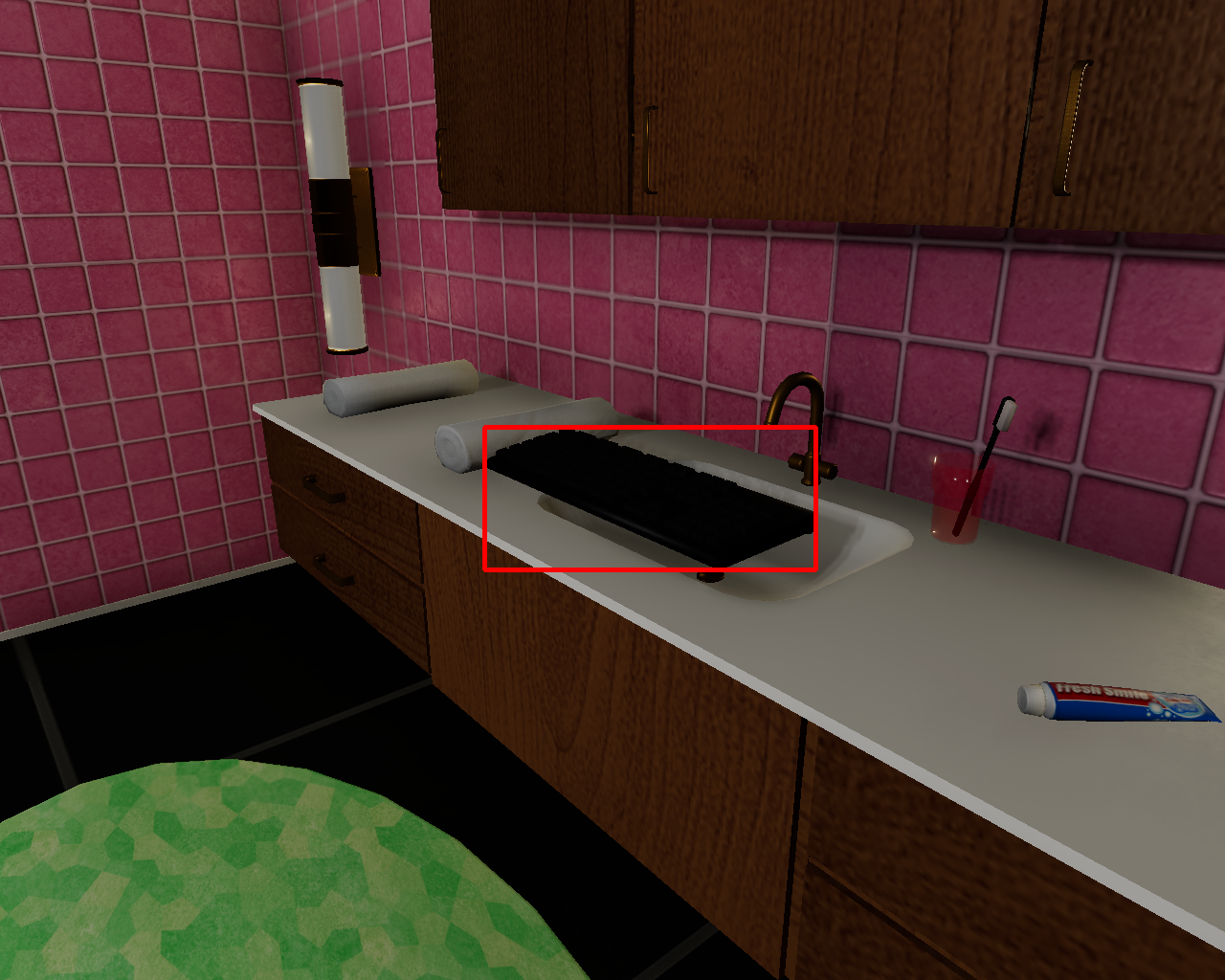}
&
\includegraphics[width=0.28\linewidth]{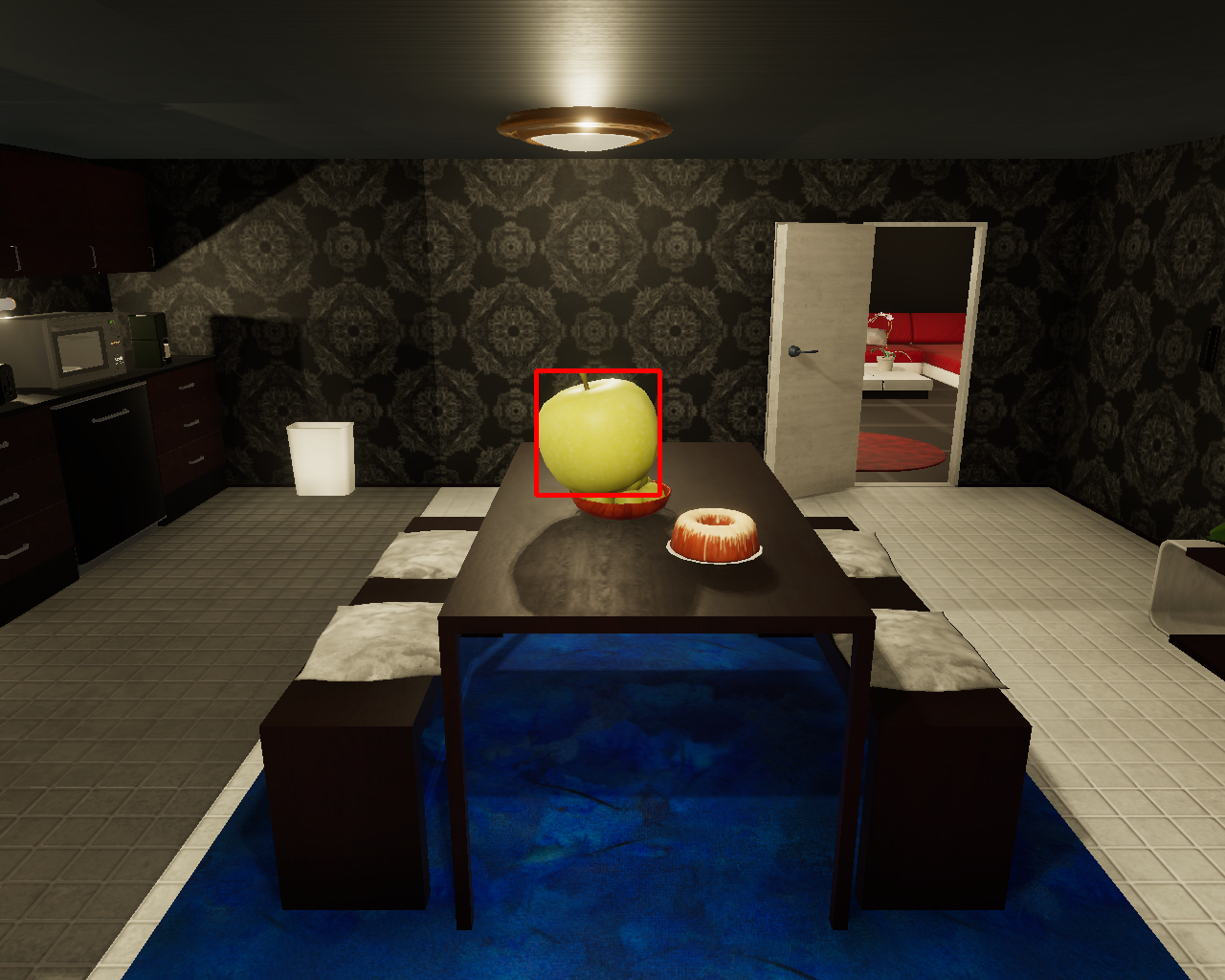}
&
\includegraphics[width=0.28\linewidth]{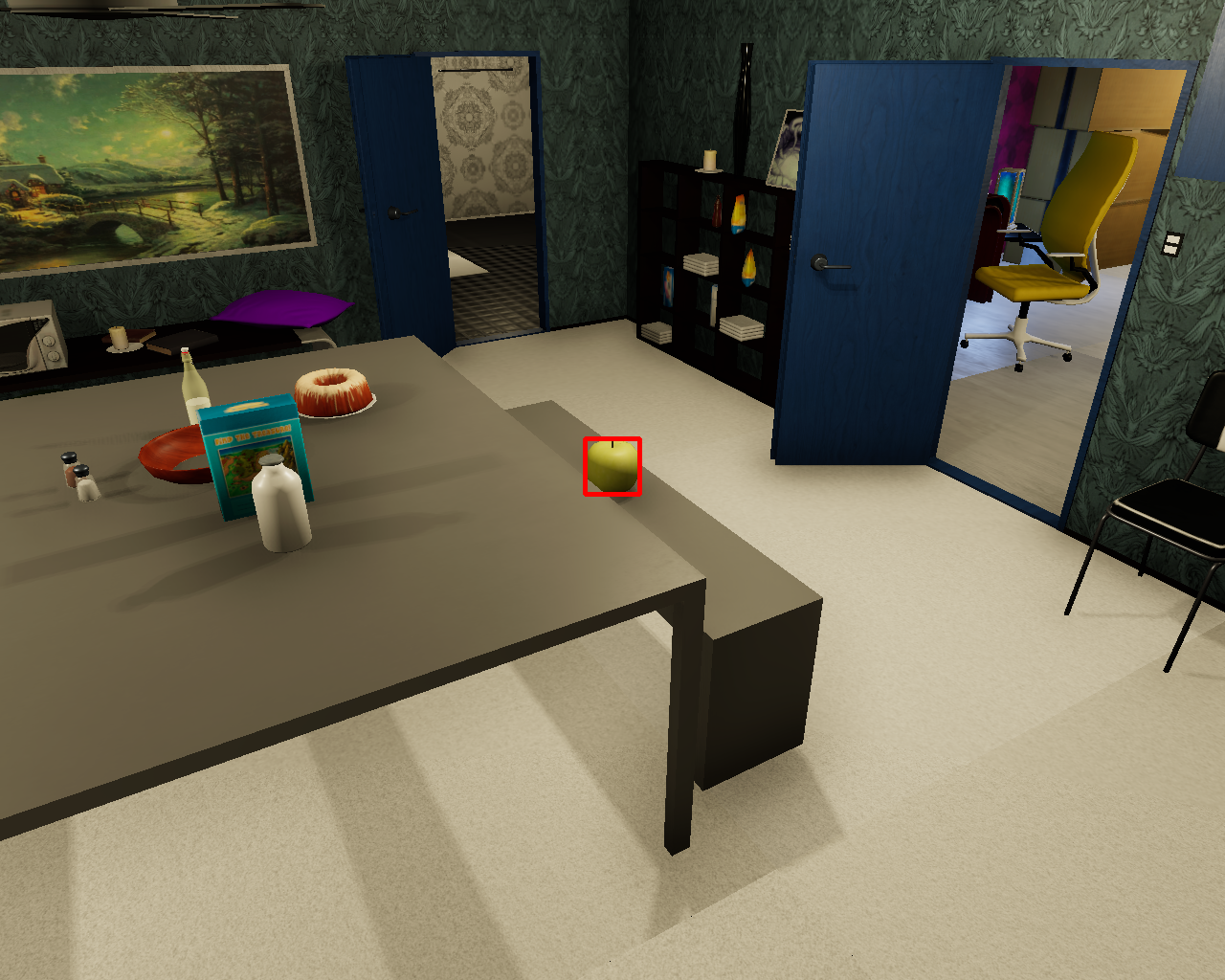}
\end{tabular}
\caption{Qualitative examples where our GCRN model correctly identifies Out-Of-Context (OOC) object instances in images. Top row presents images from our COCO-OOC dataset and bottom row presents images from the OCD dataset. The OOC objects are marked by red boxes.}
\label{fig:qualitative}
\end{figure*}

\textbf{Results on the OCD dataset.} Following the protocol in \cite{bomatter2021pigs}, we train on COCO train images and test on OOC images. As in \cite{bomatter2021pigs}, we consider the overlapping objects as OOC instances. We present the results in table~\ref{table:OCD}. Note that COCO consists of natural images and OCD consists of synthetic images. Due to the domain shift, object detection is relatively more challenging. Also context relations, such as co-occurrence and object size, in COCO can be different from those in OCD. Thus, our performance of OOC detection on OCD is lower than that of COCO-OOC.

\begin{table}
\caption{Comparison with the baselines on the OCD dataset.}
\label{table:OCD}
\centering
\begin{tabular}{lc}
\toprule
{Approach}  & {AUC score} \\
\midrule
Softmax confidence & 0.402 \\
GCRN (oracle boxes, pred labels) & 0.587
\\
GCRN (oracle boxes + labels) & 0.709
\\
\bottomrule
\end{tabular}
\end{table}

\textbf{Qualitative results.} We present the qualitative results in figure~\ref{fig:qualitative} where the top row shows the results on COCO-OOC and the bottom row shows the results on OCD. Our approach successfully detects various types of OOC objects in both natural and synthetic images. Our model is only trained on COCO dataset and can detect OOC instances in the novel OCD images.

\begin{figure}[h]
\begin{tabular}{ll}
\includegraphics[width=0.38\linewidth]{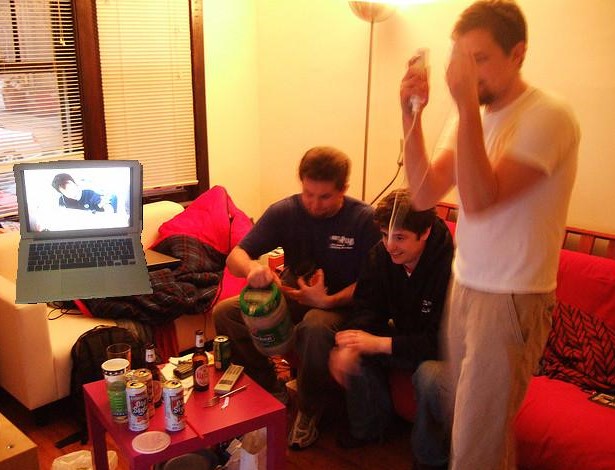}
&
\includegraphics[width=0.45\linewidth]{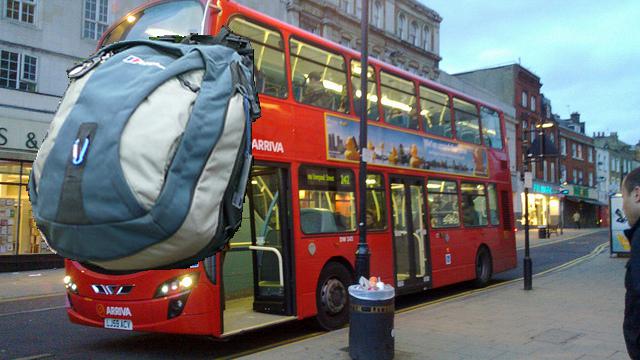}
\end{tabular}
\caption{Left: laptop is not detected as OOC, Right: backpack is not detected as OOC}
\label{fig:failure}
\end{figure}

\textbf{Failure cases.} We present two failure cases in figure~\ref{fig:failure}. In these two cases, `laptop' and `backpack' are not detected as OOC instances. Our model has observed laptops in indoor scenes with persons and indoor furniture as in-context samples. The size of the laptop is slightly larger than usual. Also, a backpack is often observed in outdoor scenes however the size of the backpack is larger than usual. The size features are often partially lost in 2D images due to projection. Having access to additional 3D shape of the objects (e.g., from RGB+Depth) can be helpful to capture context precisely.

\section{Conclusion}
\label{sec:conclusion}

We have presented an approach for detecting OOC objects in images. Detecting OOC objects is crucial to developing a reliable object detection system as detectors perform poorly on OOC objects. The proposed GCRN framework has two components: 1) RepG to learn object representation and 2) ConG to explicitly capture context for object detection. We have created a large-scale OOC dataset to evaluate our performance. We have also evaluated our approach on the recent OCD benchmark. We have considered commonly studied OOC scenarios where objects violate co-occurrence and size constraints. Our evaluation shows context cues are helpful to detect in-context objects while inconsistent context cues can hinder accurate object detection. Thus, explicit modeling of context is crucial for OOC detection, and without this OOC detection is severely affected. We have analyzed the effect of accurate object detection on the performance of OOC detection and as expected accurately localizing and identifying object labels are shown to be important for OOC detection. Finally, we present a few failure cases and propose strategies to mitigate such failures.

\clearpage
\section*{Acknowledgement}
This work was supported in part by the U.S. Army Combat Capabilities Development Command (DEVCOM) Army Research Laboratory under the IoBT REIGN Cooperative Agreement W911NF-17-2-0196, the DARPA Symbiotic Design of Cyber Physical Systems under contract FA8750-20-C-0002, and U.S. National Science Foundation grants \#1740079, \#1909696, and \#2047556. The views expressed in this paper are those of the authors and do not reflect the official policy or position of the United States Army, the United States Department of Defense, or the United States Government.

{\small
\bibliographystyle{named}
\bibliography{OOC}
}

\ifthenelse{\boolean{combined}}{
\clearpage
\begin{center}
    {\Large Supplemental Material \normalsize}
\end{center}
\beginsupplement

\section{COCO-OOC Dataset Details.}

Our COCO-OOC dataset is semi-synthetically generated which allows for more control and ensure sufficient diversity for OOC object detection. We exclusively build using the most common 80 object categories found in the COCO dataset. We hope COCO-OOC will also serve as the robustness evaluation set for the detection models trained using the MSCOCO dataset.

Co-occurrence is one of the common contextual cues that humans and vision models alike use for visual predictions. For example, if we see a grass field then we most likely expect a cow to be grazing on it or conversely if we see a beach then probably seeing a cow is not normal and thus is an out-of-context object for such scene. To generate co-occurrence violation scenarios we use 
COCO `things' ontology which categorizes COCO objects into outdoor and indoor categories. We use this categorization to determine object co-occurrence violation. In total, we consider six variations:

\begin{enumerate}[nosep]
    \item Animals in \textit{indoor} settings: bird, cat, dog, horse, sheep, cow, elephant, bear, zebra, giraffe.
    \item Outdoor objects in \textit{indoor} settings: traffic light, fire hydrant, street sign, stop sign, parking meter.
    \item Vehicles in \textit{indoor} settings: bicycle, car, motorcycle, airplane, bus, train, truck, boat.
    \item Appliances in \textit{outdoor} settings: microwave, oven, toaster, sink, refrigerator, blender.
    \item Electronic in \textit{outdoor} settings: laptop, mouse, remote, keyboard, cell phone.
    \item Food in \textit{outdoor} settings: banana, apple, sandwich, orange, broccoli, carrot, hot dog, pizza, donut, cake.
\end{enumerate}

\begin{table}[h]
\caption{OOC categories in the COCO-OOC dataset.}
\label{table:data-stats}
\centering
\begin{tabular}{lc}
\toprule
{Category}  & {\# images} \\
\midrule
Co-occurrence & 72137
\\
Size & 33899
\\
Total & 106036 
\\
\bottomrule
\end{tabular}
\end{table}

Another important contextual cue is the size of the objects. Humans have familiarity about the size of the everyday objects and can use those priors to infer about the normalcy of scenes. To generate size violation OOC scenarios, we consider the average size of objects per categories in the training set and upscale objects randomly within 2 to 5 times the original object size.

}

\end{document}